\documentclass[10pt,twocolumn,letterpaper]{article}

\usepackage{cvpr} 

\usepackage{makecell} 
\usepackage{amsmath} 
\usepackage{balance}
\usepackage{mathtools}
\usepackage{pifont}
\usepackage{multirow}
\usepackage[accsupp]{axessibility} 
\DeclarePairedDelimiter\ceil{\lceil}{\rceil}
\DeclarePairedDelimiter\floor{\lfloor}{\rfloor}

\usepackage{xcolor}
\newcommand{\xmark}[0]{\textcolor{BrickRed}{\ding{55}}}
\newcommand{\cmark}[0]{\textcolor{OliveGreen}{\ding{51}}}

\newcommand{\decreasedperf}[1]{{\color{blue}{\scriptsize    (\textbf{-#1})  }}}

\usepackage{xargs}
\usepackage[colorinlistoftodos,prependcaption,textsize=normalsize]{todonotes}
\newcommandx{\info}[2][1=]{\todo[linecolor=OliveGreen,backgroundcolor=OliveGreen!25,bordercolor=OliveGreen,#1]{#2}}
\newcommandx{\improvement}[2][1=]{\todo[linecolor=Plum,backgroundcolor=Plum!25,bordercolor=Plum,#1]{#2}}
\newcommandx{\thiswillnotshow}[2][1=]{\todo[disable,#1]{#2}}
\marginparwidth=\dimexpr \marginparwidth + 1.5cm\relax

\definecolor{cvprblue}{rgb}{0.21,0.49,0.74}
\usepackage[pagebackref,breaklinks,colorlinks,allcolors=cvprblue]{hyperref}

\title{UniSTD: Towards Unified Spatio-Temporal Learning across Diverse Disciplines}

\author{
Chen Tang$^{1}$ \quad Xinzhu Ma$^{1,2,}$\thanks{Corresponding Author} \quad Encheng Su$^{2}$ \quad Xiufeng Song$^{2,3}$ \quad Xiaohong Liu$^{3}$ \\ Wei-Hong Li$^{1,4}$ \quad Lei Bai$^{2}$ \quad Wanli Ouyang$^{1,2}$ \quad Xiangyu Yue$^{1,4,*}$ \\
{
\normalsize
$^1$MMLab, The Chinese University of Hong Kong \quad \quad  \quad $^2$Shanghai AI Lab \quad \quad \quad \quad \quad $^3$Shanghai Jiaotong University 
}\\ 
{\normalsize $^4$Shun Hing Institute of Advanced Engineering, The Chinese University of Hong Kong}\\
}

\begin{document}
\maketitle
\begin{abstract}
Traditional spatiotemporal models generally rely on task-specific architectures, which limit their generalizability and scalability across diverse tasks due to domain-specific design requirements.
In this paper, we introduce \textbf{UniSTD}, a unified Transformer-based framework for spatiotemporal modeling, which is inspired by advances in recent foundation models with the two-stage pretraining-then-adaption paradigm. 
Specifically, our work demonstrates that task-agnostic pretraining on 2D vision and vision-text datasets can build a generalizable model foundation for spatiotemporal learning, followed by specialized joint training on spatiotemporal datasets to enhance task-specific adaptability. 
To improve the learning capabilities across domains, our framework employs a rank-adaptive mixture-of-expert adaptation by using fractional interpolation to relax the discrete variables so that can be optimized in the continuous space. 
Additionally, we introduce a temporal module to incorporate temporal dynamics explicitly. 
We evaluate our approach on a large-scale dataset covering 10 tasks across 4 disciplines, demonstrating that a unified spatiotemporal model can achieve scalable, cross-task learning and support up to 10 tasks simultaneously within one model while reducing training costs in multi-domain applications. 
Code will be available at \url{https://github.com/1hunters/UniSTD}. 
\end{abstract}    
\section{Introduction}
\label{sec:intro}

In recent years, substantial progress has been made across a wide array of computer vision tasks, including image classification~\cite{tan2019efficientnet,he2016deep,dosovitskiy2020image}, object detection~\cite{zhu2020deformable,ren2015faster,he2017mask}, and more~\cite{kirillov2023segment,li2022blip,ho2020denoising,song2020denoising}. 
Among these, spatiotemporal learning—which focuses on predicting future events based on historical data—has emerged as crucial for numerous fields and practical applications, such as human motion prediction, traffic management, robotic planning, weather forecasting, \etc.
By capturing both spatial correlations and temporal dynamics, spatiotemporal learning offers a robust framework for modeling the complexities of the physical world, enabling more accurate and responsive predictions in diverse, real-world contexts. 
\begin{figure}[t]
  \centering
\includegraphics[width=1.\linewidth]{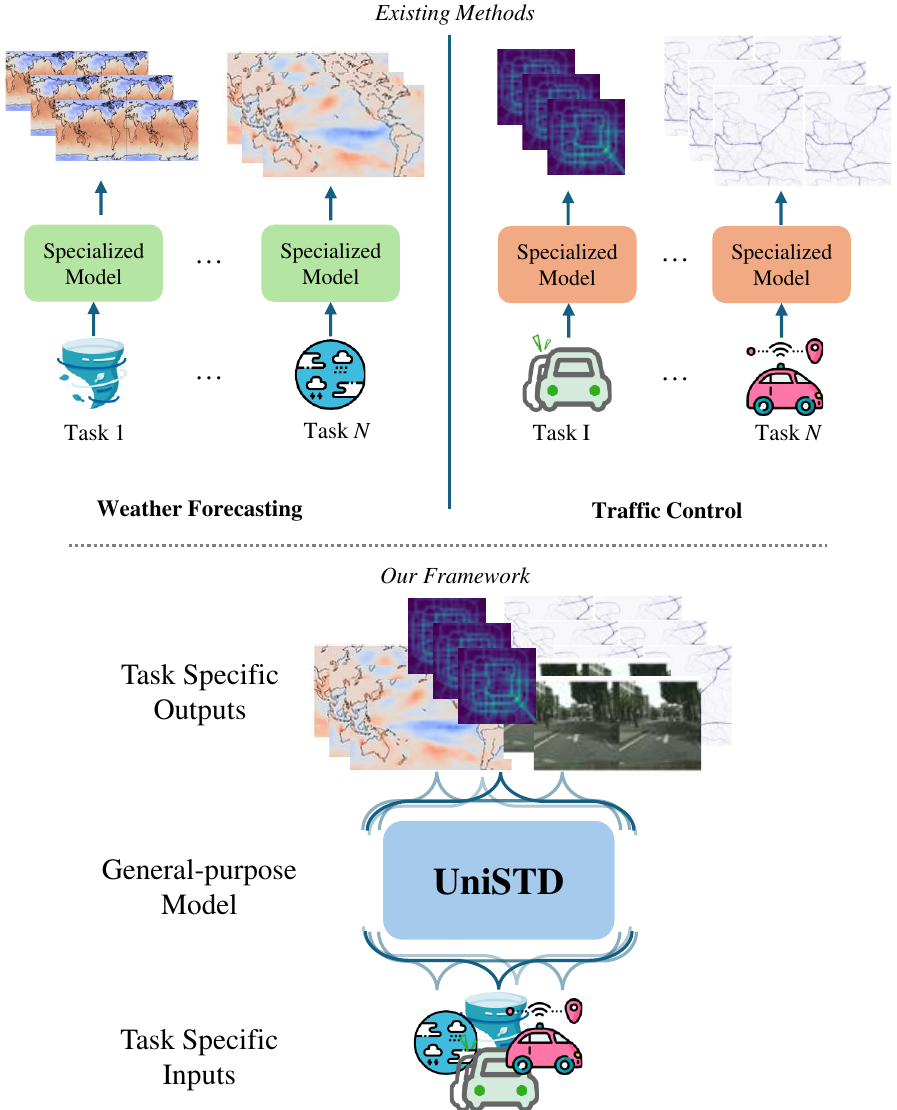}
\vspace{-0.8cm}
\caption{ 
\textbf{Top:} Existing works need specialized models for both tasks within the same disciplines (\eg, weather forecasting, traffic control). 
\textbf{Bottom:} Unified Spatio-Temporal Learning. UniSTD unifies 4 disciplines with 10 tasks under one model
 and is trained on a massive collection of datasets.}
 \vspace{-0.6cm}
\label{fig:comp_concept}

\end{figure}
While achieving meaningful progress, most of the existing methods~\cite{convlstm,predrnn,predrnn++,e3dlstm,mim,phydnet,predrnnv2,crevnet,villegas2018hierarchical,villegas2019high,babaeizadeh2021fitvid,jin2020exploring,tan2023temporal,gao2022earthformer} still resort to designing specialized architectures to learn the spatiotemporal representations for each task. 
% \xiangyu{sth. like: for different tasks?}.
% \change{Weihong: Looks like this paragraph is not end. Perhaps move while to the next paragraph.}

\emph{Recurrent-based approaches} have traditionally relied on Long Short-Term Memory (LSTM) networks~\cite{hochreiter1997long} for sequential modeling. ConvLSTM~\cite{convlstm} pioneered the application of convolutional LSTM networks for spatiotemporal predictive learning, it extends fully connected LSTM to incorporate convolutional layers. 
PredRNN~\cite{predrnn} introduced Spatiotemporal LSTM (ST-LSTM) units, enabling a unified memory pool to capture both spatial appearances and temporal dynamics. 
Furthermore, E3D-LSTM~\cite{e3dlstm} integrated 3D convolutional networks within LSTM units, facilitating robust representations that capture both short-term frame dependencies and long-term structural relationships. 
In contrast, \emph{recurrent-free approaches} have emerged to circumvent iterative temporal predictions. Instead of generating temporal sequences incrementally, these methods simultaneously produce spatial and temporal predictions. SimVP~\cite{Gao_2022_CVPR} and SimVPv2~\cite{tan2022simvp} employed a CNN backbone as a feature translator within an Encoder-Decoder framework. Earthformer~\cite{gao2022earthformer} decomposes data into cuboidal regions to parallelize spatiotemporal forecasting for earth system applications. TAU~\cite{tan2023temporal} further explores generalized architectures by introducing the Temporal Attention Unit.

While effective in eliminating recursion, existing methods primarily focus on re-engineering architectures to tackle specific tasks. 
In other words, these models are each designed for a \emph{single} task, heavily relying on task-specific domain knowledge~\cite{gao2022earthformer}. 
Consequently, they suffer from unstable performance when they are directly applied to other tasks~\cite{tan2023openstl,wang2024predbench}, limiting their generality and inevitably leading to expensive computational and memory cost.
This leads us to pose a critical, yet challenging question:
\begin{center} \textit{Can we  comprehensively address diverse spatiotemporal domains with a single, universal framework?} \end{center}
A model with such generality would offer several distinct advantages:
\emph{(i)} it allows the usage of standard network architecture (\eg, Transformer) and hence enables the integration of extensive pretrained knowledge from widely available open-source weights within the research community, 
\emph{(ii)} joint training across multiple spatiotemporal tasks could promote cross-task learning benefits, and
\emph{(iii)} It enhances scalability and operational efficiency for large-scale deployments, where simultaneous processing of tasks across various domains is required. 
The comparison between existing works and the proposed method is shown in Fig.~\ref{fig:comp_concept}. 

To unify the diverse data patterns of spatiotemporal tasks within a single model, inspired by recent advances in large language models (LLMs)~\cite{radford2018improving,radford2019language,brown2020language} and vision-language models (VLMs)~\cite{han2024onellm,liu2024visual,li2023blip}—where a well-pretrained Transformer can serve as a universal encoder for further multi-modal multi-task fine-tuning—we frame spatiotemporal learning as a two-stage optimization problem.
Since transformers has been standard blocks for various modalities, \eg, vision and text, we employ a standard transformer model to build a generalizable architecture, and further demonstrate that this choice allows us to leverage the extensive knowledge embedded in large-scale, task-agnostic pretraining on datasets such as 2D vision data and image-text pairs (\eg, OpenCLIP-ViT~\cite{ilharco_gabriel_2021_5143773} and ImageNet-ViT~\cite{dosovitskiy2020image}). 
In the second stage, which is the main focus of this paper, we perform joint training on the single model using task-specific spatiotemporal datasets for embedding domain-specific knowledge into the model, to fit the spatiotemporal tasks and improve the adaptability of the model. 

Despite the promising prospect, performing joint training on a single model is extremely challenging due to the intricate properties of diverse disciplines (\eg, weather forecasting vs. traffic control), easily triggering conflicts between disciplines and resulting in sub-optimal convergence. 
 To address this with minimal training cost, we propose a rank-adaptive Mixture-of-Experts (MoE) mechanism that dynamically optimizes low-rank adapter ranks based on task properties and interdependencies while selectively activating adapters according to input characteristics. 
To mitigate the complexity of rank optimization, we reformulate the problem using an auxiliary matrix-based approach, which reduces complexity and enables fine-grained rank adjustments. Our method achieves full differentiability by incorporating a continuous relaxation of discrete rank values, facilitating efficient optimization with minimal computational overhead. 
Additionally, to imbue models originally trained on 2D data with temporal awareness without introducing substantial computational overhead, we design a lightweight temporal module that incorporates zero-initialized projection MLP layers.
This design eliminates the need to fine-tune the transformer’s computationally intensive FFN layers, thereby enhancing temporal modeling capabilities while maintaining efficiency. 

\begin{figure*}[t]
    \centering
      \includegraphics[width=0.95\linewidth]{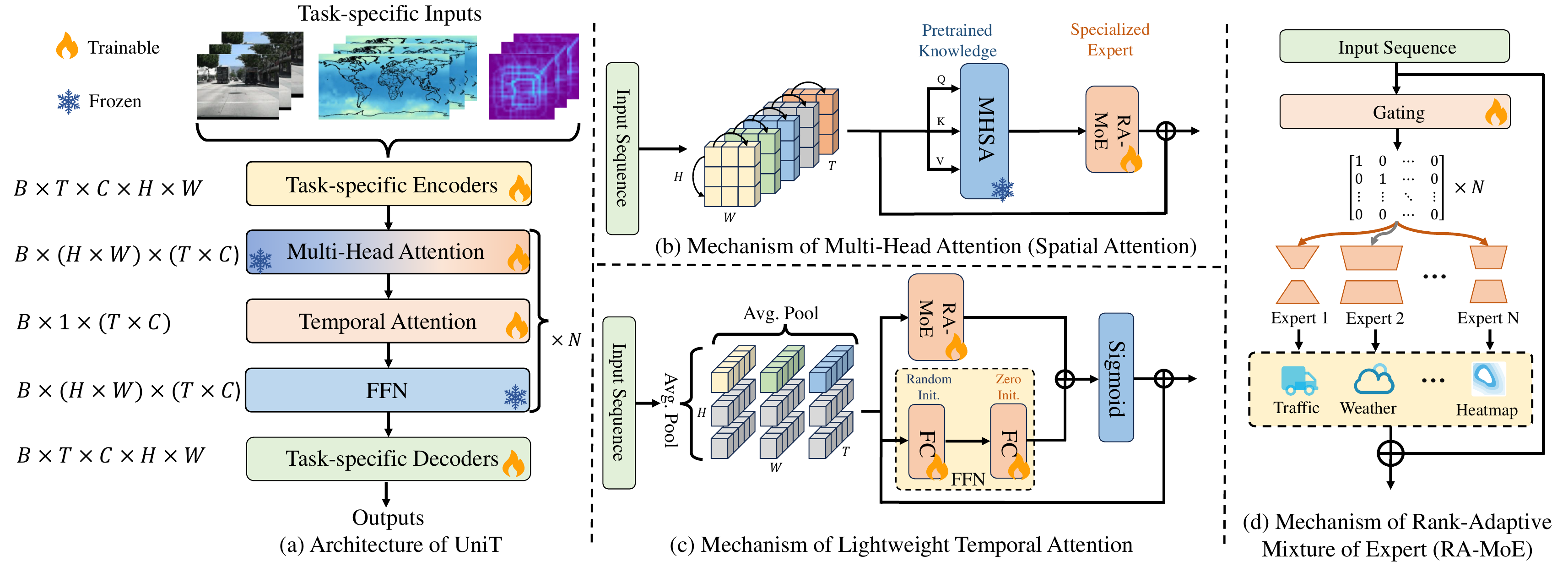}
      \vspace{-0.25cm}
      \caption{\textbf{Illustration of UniSTD.} Our method supports unified and scalable spatiotemporal learning across diverse disciplines. 
      To achieve this, we use a standard Transformer to serve as the backbone, allowing us to take advantage of the pretrained weights from large-scale task-agnostic pertaining. 
      Furthermore, to better embed the domain-specific knowledge into the model, we design a rank-adaptive MoE mechanism that dynamically adjusts the sub-architectures of model  according to the joint training process, and a lightweight temporal attention module to explicitly capture the temporal dynamics. } 
    \label{fig:main_fig}
    \vspace{-0.5cm}
  \end{figure*}

Based on OpenSTL~\cite{tan2023openstl} and PredBench~\cite{wang2024predbench}, we create a large-scale dataset encompassing 4 representative disciplines with a total of 10 tasks to support the joint training. 
Extensive experiments demonstrate the effectiveness of the proposed method, UniSTD achieves lossless performance when scaling up the number of tasks. 
Specifically, the existing methods have encountered significant performance drops at only 3 tasks jointly trained together, while at that time, UniSTD has up to 18.8 PSNR advanced compared to existing methods. The overall framework of UniSTD is shown in Fig.~\ref{fig:main_fig}. 

In summary, we make the following contributions: 

\begin{itemize}
    \item We introduce a unified framework for spatiotemporal modeling using a standard Transformer pretrained on task-agnostic datasets (\eg, OpenCLIP-ViT, ImageNet-ViT) and support specialized training on diverse spatiotemporal tasks, ensuring consistent performance, cross-task learning, and scalability with minimal reliance on domain-specific design. 
    \item 
    We decouple spatiotemporal modeling via rank-adaptive MoEs and a lightweight temporal module, enabling efficient representation of spatial and temporal dependencies. 
    \item 
    Supported by a large-scale benchmark spanning four disciplines and ten tasks, UniSTD shows impressive performance in scaling the number of tasks without performance degradation, achieving up to 18.8 PSNR improvement compared to the current methods. 
\end{itemize}

\section{Related Work}

\subsection{Spatiotemporal Predictive Learning}
\noindent \textbf{Recurrent-based methods.} 
Early works mainly focus on designing recurrent-based models for spatiotemporal predictive learning. 
ConvLSTM~\cite{convlstm} integrates convolutional networks into LSTM architectures for better spatiotemporal modeling. 
PredNet~\cite{prednet} predicts future frames using deep recurrent convolutional networks with bidirectional connections. 
PredRNN~\cite{predrnn} introduces a Spatiotemporal LSTM unit for joint spatial-temporal representation learning. 
E3D-LSTM~\cite{e3dlstm} incorporates eidetic memory in recurrent units, and Conv-TT-LSTM~\cite{su2020convolutional} leverages higher-order ConvLSTMs to combine features across time. MotionRNN~\cite{wu2021motionrnn} highlights motion trends, while LMC-Memory~\cite{lee2021video} uses memory alignment for long-term motion context. PredRNN-v2~\cite{predrnnv2} further refines PredRNN with memory decoupling loss and curriculum learning. 

\noindent \textbf{Recurrent-free methods.} 
Recurrent-free methods aim to address the inefficiency of recurrent models. 
SimVP~\cite{Gao_2022_CVPR} and SimVPv2~\cite{tan2022simvp} adopt Inception-UNet blocks to downsample the video sequences and learn spatiotemporal dynamics jointly, and then perform upsampling for prediction. While effective, convolutional methods struggle with long-term dependencies. 
MCVD~\cite{voleti2022mcvd} proposes a Masked Conditional Video Diffusion framework that uses the conditional video diffusion model based on the past-masking mechanism. 
TAU~\cite{tan2023temporal} replaces Inception-UNet with efficient attention modules, enabling parallelization and improved long-term temporal learning. 
Furthermore, WaST~\cite{nie2024wavelet} adopts a 3D discrete wavelet transform module to extract low and high-frequency components jointly for better long-term dependency modeling. 
EarthFormer~\cite{gao2022earthformer} leverages the space-time attention block for earth system forecasting, which decomposes the input into cuboids and then applies cuboid-level attention in parallel. 
Despite their promising performance, existing methods primarily focus on designing specialized architectures to tackle
single spatiotemporal tasks. In this paper, we propose a general and scalable model based on a pure Transformer architecture to enable unified spatiotemporal learning.

\subsection{General-purpose Model} 
Numerous efforts~\cite{caruana1997multitask,ruder2017overview,vandenhende2021multi} have been made to develop a general-purpose model capable of handling diverse tasks in a unified framework. 
Early works in the field of natural language processing (NLP)~\cite{kaiser2017one,brown2020language,jaegle2021perceiver} and computer vision (CV)~\cite{misra2016cross,kokkinos2017ubernet,zamir2020robust,li2022learning,li2024universal} demonstrated the feasibility of designing a unified framework to handle cross-task prediction or generation. 
ExT5~\cite{aribandi2021ext5}, UniT~\cite{hu2021unit}, and OFA~\cite{wang2022ofa} further demonstrate the large-scale multi-task joint training is of importance to the performance. 
UniHead~\cite{liang2022unifying}, UniHCP~\cite{ci2023unihcp}, and UViM~\cite{kolesnikov2022uvim} leverage unified architectures (typically Transformer) to learn the shared representations for multiple perception vision tasks. 
Emu~\cite{sun2023generative}, SEED~\cite{ge2024seed}, and Mini-gemini~\cite{li2024mini} leverage both vision-language models and diffusion models to achieve text and visual generation. 
However, these methods do not focus on spatiotemporal learning, which has more diversity between tasks. 
\section{Method}
\subsection{Unified Modeling with Transformer} 
\noindent 
\textbf{Preliminary.} 
Spatiotemporal learning aims to infer the future frames using the previous ones. 
For a model that simultaneously supports $M$ spatiotemporal tasks, each task $i$ takes a sequence of historical information and predicts the future sequence. 
More formally, 
$ \boldsymbol{X}^{(i)}_{T_{i}}=\{ \boldsymbol{x}_j \}_{t - T_{i} + 1}^{t}$ at time $t$ with the past $T$ frames, the output is $\boldsymbol{Y}^{(i)}_{{T'}_{i}} = \{ \boldsymbol{y}_j \}_{t}^{t + {T'}_{i}}$ that contains the next $T'$ frames, where $\boldsymbol{y}_j \in \mathbb{R}^{C_{i} \times H_{i} \times W_{i}}$ is an input with channels $C_i$, height $H_i$, and width $W_i$. 
Overall, the model is optimized by:
\begin{equation}
    \Theta^* = \arg\min_{\Theta} \mathcal{L}_{\text{MSE}}(\mathcal{F}_{\Theta}(\boldsymbol{X}), \boldsymbol{Y}),
\end{equation}
where $\Theta$ is the parameters of the model, $\boldsymbol{X} = \{\boldsymbol{X}^{(i)}_{T_{i}}\}_{i=1}^{M}$, $\mathcal{L}_{\text{MSE}}$ denotes the mean squared error loss function. $\mathcal{F}_{\Theta}$ denotes the standard Transformer model \citep{vaswani2017attention} that alternates between Multi-Head Attention (MSA) and Feed-Forward Networks (FFN). The attention mechanism for each head is defined as:
\begin{align}
\texttt{Attention}(\mathbf{Q}, \mathbf{K}, \mathbf{V}) = \texttt{Softmax}\left(\frac{\mathbf{Q}\mathbf{K}^\top}{\sqrt{L}}\right) \mathbf{V},
\end{align}
where in self-attention, the queries \(\mathbf{Q}\), keys \(\mathbf{K}\), and values \(\mathbf{V}\) are linear projections of the input \(\mathbf{x}\), represented as \(\mathbf{Q} = \mathbf{x}\mathbf{W}^{Q}\), \(\mathbf{K} = \mathbf{x}\mathbf{W}^{K}\), and \(\mathbf{V} = \mathbf{x}\mathbf{W}^{V}\), with \(\mathbf{x} \in \mathbb{R}^{N \times L}; \; \text{and} \; \mathbf{Q}, \mathbf{K}, \mathbf{V} \in \mathbb{R}^{L \times L} \), where $N$ denotes the number of tokens, $L$ is the hidden embedded dimension of Transformer. 
A linear layer then projects the output of self-attention. 
The FFN then processes each position in the sequence by applying two linear transformations: 
\begin{align}
\mathbf{Y} &= \texttt{FFN}(\texttt{LN}(\mathbf{x})) + \mathbf{x}.  
\end{align}

\noindent \textbf{Encoder and Decoder architectures. } 
In the encoder, we employ a canonical design in spatiotemporal representation learning~\cite{tan2023temporal}, which incorporates a series of \texttt{2DConv-GroupNorm-SiLU} layers to progressively downsample the spatial dimensions of the input. 
Given an input tensor with the shape $(B \times T_i) \times C_i \times H_i \times W_i$, where $B$ denotes the batch size and $T_i$ is folded into the batch dimension, the encoder produces an output with reduced spatial dimensions $H_i'$ and $W_i'$, and an updated channel dimension $C_i'$, \ie, the shape is $(B \times T_i) \times C_i' \times H_i' \times W_i'$. 
Then, it is reshaped into $B \times N \times L$ to serve as input to the Transformer, where $N = H_i' \times W_i'$ represents the spatial token count, and $L = T_i \times C_i'$ corresponds to the feature dimensionality of each token. 

In the decoder, we apply an inverse process to that of the encoder. The output of the Transformer, initially in the shape $B \times N \times L$, is first reshaped to match the input format of the Transformer, specifically $(B \times T_i) \times C_i' \times H_i' \times W_i'$. It is then upsampled along the spatial dimension using transposed convolution. 
\\ 
\noindent \textbf{Position Encoding}. We adopt sinusoidal position encoding (SPE) generated with absolute coordinates for each patch, instead of learnable position embeddings used in existing Transformer models (\eg, ViT, GPT). \\ 
\noindent \textbf{Initialization. } 
As discussed previously, pretrained Transformers provide strong modality adaption capability for various tasks (\eg, 2D vision, vision-language understanding, \etc), we initialize the Transformer backbone (except for the task-specific encoder and decoder) with the pretrained weights. 
\\

\subsection{Specialized Training} 
Although the standard Transformer allows us to use the weights on task-agnostic pertaining, we still need to perform specialized training for task-specific adaptability. 
However, with the remarkable scalability of modern Transformer models, the size of pre-trained models is increasing exponentially to achieve superior performance. 
As a result, the storage cost of the full training paradigm becomes prohibitive in multi-task scenarios~\cite{hulora,jie2023revisiting}. 
Therefore, we implement the specialized training with low-rank adaptation (LoRA)~\cite{hulora}, which is a parameter-efficient finetuning technique that can reduce the number of fine-tuning parameters and memory usage. Specifically, 
given an input $\mathbf{x} \in \mathbb{R}^{N \times L}$, and original frozen weights $\mathbf{W}^{Q/K/V}$ in the multi-head self-attention layers, the low-rank adapter consist of two low-rank trainable matrices $\mathbf{A} \in \mathbb{R}^{L \times r'}$ and $\mathbf{B} \in \mathbb{R}^{r' \times  L }$, where $r' \ll L$: 
\begin{equation} 
    \mathbf{Y}_{r'}^{Q/K/V} = \mathbf{x} \underbrace{\mathbf{W}^{Q/K/V}}_{\text{frozen}} + \alpha \cdot \mathbf{x} \underbrace{\mathbf{A}_{r'} \mathbf{B}_{r'}}_{\text{trainable}}
    \label{equ:lora}
\end{equation} 
\begin{figure}[t]
    \includegraphics[width=1.0\linewidth]{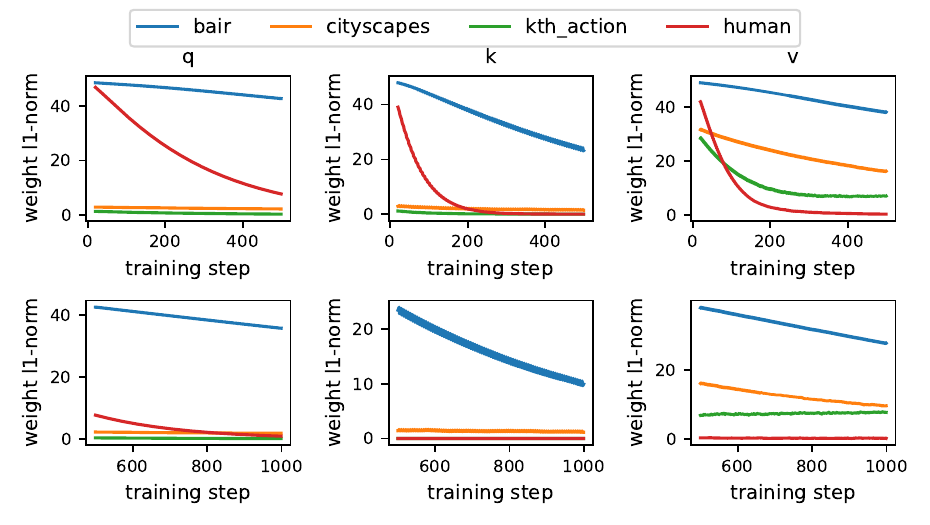}
    \vspace{-0.35cm}
    \caption{Weight updating patterns of task-specialized experts during optimization. We demonstrate the L1-norm of queries (Q), keys (K), and values (V) as a measurement for the optimal rank of each expert. }
    \label{fig:qkv_norm}
    \vspace{-0.5cm}
\end{figure} 

\noindent\textbf{Rank-adaptive MoE. } 
As shown in Fig.~\ref{fig:qkv_norm}, we have observed that tasks have quite discrepant update patterns of the weight, this suggests making all tasks share a \emph{single} low-rank adapter is easy to yield sub-optimal performance. 
Therefore, to effectively balance multi-spatiotemporal tasks spanning diverse disciplines with highly distinct characteristics (\eg, weather, traffic, \etc), we employ the mixture-of-experts paradigm. 
Specifically, a dynamic router $\mathcal{G}(\cdot)$ is utilized to adaptively assign weights $\mathrm{w}$ to low-rank adapters based on the requirements of the specific input task/discipline. 
For low-rank adapter in $l$-th layer, \cref{equ:lora} becomes: 
\begin{equation} 
    \mathbf{Y}^{Q/K/V}_{{r'}} = \mathbf{x} \underbrace{\mathbf{W}^{Q/K/V}}_{\text{frozen}} + \alpha \cdot  \sum_{i=1}^{N}  \mathbf{x} \underbrace{\mathcal{G}^{(i)}(\mathbf{x}) \mathbf{A}_{r'}^{(i)} \mathbf{B}_{r'}^{(i)}}_{\text{trainable}}, 
    \label{equ:moe_lora}
\end{equation} 
where $\mathcal{G}^{(i)}(\cdot)$ denotes the $i$-th component. 

Fig.~\ref{fig:qkv_norm} illustrates that $\mathbf{Q}$, $\mathbf{K}$, and $\mathbf{V}$ exhibit task-specific weight update patterns, implying that \emph{the rank $r'$ of each expert should be dynamically adjusted based on task properties and interdependencies}, rather than being fixed or pre-defined. However, direct optimization of the discrete rank $r'$ poses a combinatorial challenge. 
Considering a low-rank MoE model of $10$ layers, with each layer containing $5$ experts, and $4$ potential rank configurations for each expert, the dimensionality of the search space dramatically explodes to $4^{5 \times 10}$. 
This exponential growth is computationally prohibitive, as each expert requires independent storage for its rank configurations. Additionally, an exhaustive brute-force search over this space is computationally infeasible. 

In addressing this issue, we first convert the original form to an equivalent form from the perspective of the block matrix and use a new identity matrix $\mathbf{I}_{r-1, r}$, where $r$ is an arbitrary rank of $\mathbf{A}$ and $\mathbf{B}$ to serve as the maximum of the rank: 
\begin{align}
     \Big( \mathbf{A} \mathbf{B} \Big)_{r-1} & =  \mathbf{A} \mathbf{I}_{r-1, r} \mathbf{B} = \begin{bmatrix}
         \mathbf{a}_1  & \cdots  &   \mathbf{a}_r
        \end{bmatrix} \, \mathbf{I}_{r-1, r} \, \begin{bmatrix}
         \mathbf{b}_1 \\
         \vdots \\
         \mathbf{b}_r
        \end{bmatrix}, \nonumber \\ 
    % & = \sum_{i=1}^{r-1} \mathbf{b}_i \mathbf{a}_i  \\
    & \text{where} \quad \mathbf{I}_{r-1, r} =   
    \scriptsize{
    \begin{bmatrix} 
    1 & 0 &  \cdots & 0\\ 
    0 & 1 &  \cdots & 0 \\ 
    0 & 0 &  \cdots & 0 \\ 
    \vdots & \vdots &  \ddots & \vdots 
    \\ 0 &  0 & \cdots & 0\end{bmatrix}_{r \times r}}. 
    \label{equ:identity_m_sol}
\end{align}
Here, we omit the index of expert and the rank $r$ for simplicity. 
To be precise, $\mathbf{I}_{r-1, r}$ represents a modified $r \times r$ identity matrix, where the first $r-1$ diagonal elements are $1$, and the remaining $r - (r-1)$ diagonal elements are $0$. Utilizing block matrix multiplication, it can be demonstrated that the rank of $\big( \mathbf{B} \mathbf{A} \big)_{r-1}$ is reduced to $r-1$. For any $1 \leq k < r$, it is trivial to build a rank $r-k$ solution according to \cref{equ:identity_m_sol}. 
Unlike the original form that requires storing $K$ copies for one expert, this new form allows for precise control of the rank by constraining the number of nonzero elements on the main diagonal of the original identity matrix.

However, the new form still explicitly needs a discrete $k$ to acquire the rank, which is non-differentiable for optimization. 
Then, we consider using relaxation to make it continuous by leveraging the fact that any continuous value is between its two adjacent discrete values. 
Specifically, we define the continuous rank from the integer domain as follows: 
\begin{align}
    & f_{r}(\mathbf{x})  \triangleq \big(\ceil{r} - r \big) g_{\ceil{r}} \left(\mathbf{x} \right) + \big(r - \floor{r} \big) g_{\floor{r}} \left(\mathbf{x} \right),  \nonumber \\  \nonumber \\
    &  \text{where} \quad g_{*}(\mathbf{x}) = 
    \begin{cases} 
    \mathbf{x}\mathbf{A} \mathbf{I}_{n} \Big[:\floor{r} \Big] \mathbf{B} & \text{if * is  } \floor{r} \\ \\
    \mathbf{x}\mathbf{A} \mathbf{I}_{n} \Big[:\ceil{r} \Big] \mathbf{B} & \text{if * is  } \ceil{r} 
    \end{cases} 
    \label{equ:continuous_rank}
\end{align} 
It is clear that if the output of $f_{r}(\mathbf{x})$ is bounded by the two integer values of the rank $r$, i.e., $ g_{\floor{r}} \left(\mathbf{x} \right) \leq f_{r} (\mathbf{x}) \leq g_{\ceil{r}} \left(\mathbf{x} \right)$. 
Substitute \cref{equ:continuous_rank} into \cref{equ:moe_lora}, we have: 
\begin{equation}
    \mathbf{Y}^{Q/K/V}_{{r}} = \mathbf{x} \mathbf{W}^{Q/K/V} + \alpha \cdot  \sum_{i=1}^{N}  \mathcal{G}^{(i)}(\mathbf{x}) f^{(i)}_r(\mathbf{x}).  
\end{equation}
Accordingly, the gradient of $r$ is hereby defined by the difference of the upper bound and lower bound: 
\begin{equation}
    \frac{\partial f_{r}(\mathbf{x})}{\partial r} = \gamma \cdot \Big( g_{\ceil{r}} - g_{\floor{r}} \Big), 
\end{equation} 
where $\gamma$ is a hyper-parameter to compensate the approximation error.

Once the rank $r$ becomes continuous, another benifit is that we can directly restrict the additional trainable parameters for controlling the storage overhead, which is crucial otherwise the rank may become as large as possible. 
Specifically, we can control the expected size using the L1 distance to directly optimize the summed $r$ of each expert. 
Therefore, the overall training objective is: 
\begin{equation}
    \mathcal{L} = \mathcal{L}_{\text{MSE}} + \beta \cdot \Big| C - \sum_{i=1}^N r_i \Big|, 
\end{equation} 
where $\beta$ is the hyper-parameter to weight the MSE loss and L1 loss, and $C$ is the target size of summed LoRA modules. 
We use the first 10 epochs for optimizing the rank with rank-adaptive MoE, after that, we use the $\texttt{round}$ operator to discretize the rank for the remaining training. 
\\
\noindent \textbf{Lightweight Temproal Attention. } 
We propose a lightweight temporal attention module designed to explicitly capture temporal dependencies after the self-attention layer. 
Since the FFN layers in the Transformer serve as mixers along the final dimension of the input sequence~\cite{yu2022metaformer,yu2023metaformer}, which corresponds to the temporal dimension in our framework, we introduce a new FFN layer consisting of a projection-down layer with the ratio of 6, a nonlinearity, and a corresponding projection-up layer to achieve efficient modeling. Moreover, we integrate a 1D rank-adaptive MoE (RA-MoE) layer, empowering adaptive capacity for multi-discipline learning. The resulting output is then added to the original sequence $\mathbf{x}$.

Furthermore, in contrast to prior approaches such as the SE module~\cite{hu2018squeeze} and TAU module~\cite{tan2023temporal}, which initialize their parameters randomly, we employ a zero-initialization strategy for the second MLP layer within the newly introduced FFN to preserve the pre-trained state of the original Transformer FFN layers and enable more stable optimization. 
More formally, for the input sequence $\mathbf{x} \in \mathbb{R}^{N \times L}$, Let $\mathbf{x'} = \texttt{AVGPOOL}(\mathbf{x})$, where $\mathbf{x'} \in \mathbb{R}^{1 \times L}$: 
\begin{align}
    \mathbf{o} &= \texttt{FFN}(\mathbf{x'}) + \texttt{RA-MoE}_{1D}(\mathbf{x'}) \nonumber \\
    \mathbf{x} &= \mathbf{x} + \texttt{SIGMOID} (\mathbf{o}).
\end{align}

\section{Experiments}
\subsection{Experimental Setup} 
\paragraph{Datasets.} We quantitatively evaluate our model on the following disciplines with both synthetic and real-world scenarios datasets: 
\textbf{(i) Traffic Control}, including TaxiBJ~\cite{zhang2018predicting}, and Traffic4Cast ~\cite{eichenberger2022traffic4cast} datasets. 
\textbf{(ii) Trajectory Prediction and Robot Action Planning}, 
including Moving MNIST~\cite{srivastava2015unsupervised}, BAIR~\cite{ebert2017self}, Human3.1M~\cite{ionescu2013human3}, and KTH Action~\cite{schuldt2004recognizing} datasets. 
\textbf{(iii) Driving Scene Prediction}, including Cityscapes~\cite{cordts2016cityscapes} and KITTI~\cite{geiger2013vision} datasets. 
\textbf{(iv) Weather Forecasting}, including SEVIR~\cite{veillette2020sevir} and ENSO~\cite{ICARENSO} datasets. 
We summarize the statistics of the above datasets in Tab.~\ref{tab:dataset_statistics}, including the number of training samples $N_{train}$ and the number of testing samples $N_{test}$. \\
\noindent \textbf{Evaluation Metrics.} 
We employ Structure Similarity Index Measure (SSIM), and Peak Signal to Noise Ratio (PSNR) to evaluate the quality of predictions. SSIM measures the similarity of structural information within the spatial neighborhoods, and PSNR is an expression for the ratio between the maximum possible power of a signal and the power of distorted noise. 
Notably, for weather forecast tasks, we adopt the Critical Success Index (CSI) for the SEVIR dataset and the three-month-moving-averaged Nino3.4 index (NINO) for ENSO dataset.\\
\noindent \textbf{Implementation Details.}  
UniSTD is implemented using the PyTorch framework. 
We train all the models for 90 epochs with a mini-batch size of 16, employing the AdamW optimizer configured with a learning rate of 0.01 and weight decay of 0.05. 
We utilize the standard Vision Transformer architecture (ViT base), comprising 12 layers, an embedding dimension ($L$) of 768, and a MLP expansion ratio of 4. Each self-attention layer contains 12 attention heads. If the encoder output dimensionality does not match the embedded dimension required by the Transformer, an additional linear projection layer is employed to align the dimensions accordingly.
Rank-adaptive MoE modules are integrated into the self-attention layers for the query ($\mathbf{Q}$), key ($\mathbf{K}$), value ($\mathbf{V}$), and projection (Proj) matrices, as well as within the newly introduced temporal attention layer. 
The number of experts is set to 6 for the $\mathbf{Q}$, $\mathbf{K}$, and $\mathbf{V}$ matrices, while the Proj matrices and temporal attention layer utilize 2 experts each. The initial rank for each MoE layer is fixed at 4.5, and the hyperparameter $\beta$ is set to 1.

\begin{table}[t]
    \centering
    \small
    \caption{Statistics of the datasets used in experiments. Each dataset has different input/output lengths and dimensions.}
    \begin{tabular}{lcccccc}
        \toprule
        Dataset & $N_{\text{train}}$ & $N_{\text{test}}$ & (C, H, W) & $T_{\text{input}}$ & $T_{\text{output}}$ \\
        \midrule
        KITTI         & 9209  & 2198  & (3, 128, 160) & 10 & 10  \\
        BAIR          & 38937 & 256   & (3, 64, 64)   & 2  & 10  \\
        Cityscapes    & 8925  & 1525  & (3, 128, 128) & 2  & 5   \\
        TaxiBJ        & 20461 & 500   & (2, 32, 32)   & 4  & 4   \\
        SEVIR         & 44760 & 12144 & (1, 384, 384) & 13 & 12  \\
        MMNIST        & 10000 & 10000 & (1, 64, 64)   & 10 & 10  \\
        KTH Action    & 8488  & 5041  & (1, 128, 128) & 10 & 10  \\
        Human         & 73404 & 8582  & (3, 256, 256) & 4  & 4   \\
        Traffic4Cast  & 35840 & 4508  & (8, 128, 112) & 9  & 3   \\
        ENSO          & 52350 & 1679  & (1, 48, 48)   & 12 & 14  \\
        \bottomrule
    \end{tabular}
    \label{tab:dataset_statistics}
\end{table}

\begin{table*}[t]
  \centering
  \footnotesize
  \caption{Main results of UnSTD. Both baselines and UnSTD are trained on the specific spatiotemporal datasets with one model. SimVP$_{\textbf{5 Tasks}}$ indicates that one SimVP model is trained on 3 datasets jointly. PSNR\&SSIM: higher is better. }
  \setlength\tabcolsep{1.5pt} 
  \renewcommand\arraystretch{1.3}
    
    \begin{tabular}{l|rrcc|ccrrrrcc|rrcc|cc}
    \toprule
          & \multicolumn{4}{c|}{\textbf{Traffic Control}} & \multicolumn{8}{c|}{\textbf{Trajectory Prediction and Robot Action Planning}} & \multicolumn{4}{c|}{\textbf{Driving Scene Prediction}} & \multicolumn{2}{c}{\textbf{Weather Forecasting}} \\
    \multicolumn{1}{c|}{\multirow{2}[1]{*}{Model}} & \multicolumn{2}{c}{\rotatebox[origin=c]{45}{TaxiBJ}} & \multicolumn{2}{c|}{\rotatebox[origin=c]{45}{Traffic4Cast}} & \multicolumn{2}{c}{\rotatebox[origin=c]{45}{MMNIST}} & \multicolumn{2}{c}{\rotatebox[origin=c]{45}{BAIR}} & \multicolumn{2}{c}{\rotatebox[origin=c]{45}{Human3.1M}} & \multicolumn{2}{c|}{\rotatebox[origin=c]{45}{KTH}} & \multicolumn{2}{c}{\rotatebox[origin=c]{45}{Cityscapes}} & \multicolumn{2}{c|}{\rotatebox[origin=c]{45}{KITTI}} & \multicolumn{1}{c}{\rotatebox[origin=c]{45}{SEVIR}} & \multicolumn{1}{c}{\rotatebox[origin=c]{45}{ENSO}} \\
          & \multicolumn{1}{l}{PSNR} & \multicolumn{1}{l}{SSIM} & \multicolumn{1}{l}{PSNR} & \multicolumn{1}{l|}{SSIM} & \multicolumn{1}{l}{PSNR} & \multicolumn{1}{l}{SSIM} & \multicolumn{1}{l}{PSNR} & \multicolumn{1}{l}{SSIM} & \multicolumn{1}{l}{PSNR} & \multicolumn{1}{l}{SSIM} & \multicolumn{1}{l}{PSNR} & \multicolumn{1}{l|}{SSIM} & \multicolumn{1}{l}{PSNR} & \multicolumn{1}{l}{SSIM} & \multicolumn{1}{l}{PSNR}  & \multicolumn{1}{l|}{PSNR} & \multicolumn{1}{c}{CSI}  & \multicolumn{1}{c}{NINO} \\
    \midrule
    SimVPv1$_{\textbf{3 Tasks}}$ & \multicolumn{2}{c}{-}  & 23.5 & 0.05 & \multicolumn{2}{c}{-} & \multicolumn{2}{c}{-}    & \multicolumn{2}{c}{-}    & 9.8 & 0.32 & \multicolumn{2}{c}{-} & 14.3 & 0.27 & - & - \\ 
    SimVPv2$_{\textbf{3 Tasks}}$ & 24.6  & 0.56  & \multicolumn{2}{c|}{-} & \multicolumn{2}{c}{-} & 15.7  & 0.53  & 22.4  & 0.88  & \multicolumn{2}{c|}{-} & \multicolumn{2}{c}{-} & \multicolumn{2}{c|}{-} & - & - \\ 
    SimVPv2$_{\textbf{3 Tasks}}$ & \multicolumn{2}{c}{-}  & 23.7 & 0.08 & \multicolumn{2}{c}{-} & \multicolumn{2}{c}{-}    & \multicolumn{2}{c}{-}    & 17.0 & 0.56 & \multicolumn{2}{c}{-} & 12.4 & 0.18 & - & - \\ 
    SimVPv2$_{\textbf{5 Tasks}}$ & 20.4  & 0.43  & \multicolumn{2}{c|}{-} & \multicolumn{2}{c}{-} & 12.1  & 0.34  & 17.6  &  0.73  & \multicolumn{2}{c|}{-} & 14.7  & 0.46  & \multicolumn{2}{c|}{-} &       0.32  & - \\ 
    TAU$_{\textbf{3 Tasks}}$   & 28.2    & 0.72  & \multicolumn{2}{c|}{-} & \multicolumn{2}{c}{-} & 16.5  & 0.57   & 21.3  & 0.84  & \multicolumn{2}{c|}{-} & \multicolumn{2}{c}{-} & \multicolumn{2}{c|}{-} & - & -\\
    TAU$_{\textbf{3 Tasks}}$   & 30.5  & 0.78  & \multicolumn{2}{c|}{-} & \multicolumn{2}{c}{-} & \multicolumn{2}{c}{-} & 22.4 & 0.86  & \multicolumn{2}{c|}{-} & 18.9 & 0.59   & \multicolumn{2}{c|}{-} & - &- \\
    TAU$_{\textbf{3 Tasks}}$   & \multicolumn{2}{c}{-} & \multicolumn{2}{c|}{-} & \multicolumn{2}{c}{-} & 16.7  & 0.59  & 21.2  & 0.84  & \multicolumn{2}{c|}{-} & 18.6  & 0.57  & \multicolumn{2}{c|}{-} &  - & - \\
    TAU$_{\textbf{3 Tasks}}$ & \multicolumn{2}{c}{-}  & 23.8 & 0.08 & \multicolumn{2}{c}{-} & \multicolumn{2}{c}{-}    & \multicolumn{2}{c}{-}    & 17.4 & 0.31 & \multicolumn{2}{c}{-} & 13.5 & 0.17 & - & - \\ 
    Eformer$_{\textbf{3 Tasks}}$ & 23.2 & 0.58  & \multicolumn{2}{c|}{-} & \multicolumn{2}{c}{-} & \multicolumn{2}{c}{-}  & 15.2 & 0.6   & \multicolumn{2}{c|}{-} & 19.3  & 0.63  & \multicolumn{2}{c|}{-} & - & - \\
    Eformer$_{\textbf{3 Tasks}}$ & 23.3  & 0.60  & \multicolumn{2}{c|}{-} & \multicolumn{2}{c}{-} & 12.9 & 0.38  & \multicolumn{2}{c}{-} & \multicolumn{2}{c|}{-} & 17.2  & 0.57  & \multicolumn{2}{c|}{-} & - & - \\
    Eformer$_{\textbf{3 Tasks}}$ & \multicolumn{2}{c}{-}   & 24.0 & 0.83 & \multicolumn{2}{c}{-} & \multicolumn{2}{c}{-} & \multicolumn{2}{c}{-}  & 16.8 & 0.74 &   \multicolumn{2}{c}{-}  & 10.9 & 0.32 & - & - \\
    Eformer$_{\textbf{5 Tasks}}$ & 20.8 & 0.41  & \multicolumn{2}{c|}{-} & \multicolumn{2}{c}{-} & 12.0 & 0.32  & 15.0 & 0.69  & \multicolumn{2}{c|}{-} & 14.1 & 0.41  & \multicolumn{2}{c|}{-} &  0.30   & - \\
    \textbf{Ours}$_{\textbf{10 Tasks}}$ & \textbf{39.6} & \textbf{0.98}  & \textbf{30.6} &  \textbf{0.92} & \textbf{20.5} & \textbf{0.90}  & \textbf{20.3} & \textbf{0.86}  & \textbf{33.2} & \textbf{0.98} & \textbf{28.4} & \textbf{0.92}  & \textbf{27.4} & \textbf{0.89} &  \textbf{17.2} & \textbf{0.61} &    \textbf{0.41}   & \textbf{0.72} \\
    \bottomrule
    \end{tabular}%
  \label{tab:main_results}%
\end{table*}% 

\begin{table}[t]
  \centering
  \caption{Task-wise comparison of our unified model and the single task baselines. }
  \renewcommand\arraystretch{1.3}
    \begin{tabular}{c|lcc}
    \toprule
    \multicolumn{1}{c}{Task} & Model & \multicolumn{1}{c}{PSNR} & \multicolumn{1}{c}{SSIM} \\
    \midrule
    \multirow{5}[2]{*}{\rotatebox[origin=c]{90}{TaxiBJ}} 
    & Ours &   39.6    & 0.9825 \\
    & TAU   &   39.3 \decreasedperf{0.3}   & 0.9813 \decreasedperf{0.0012} \\
          & SimVP &   39.2 \decreasedperf{0.4}   & 0.9820  \decreasedperf{0.0005} \\
          & SimVPv2 &  39.2  \decreasedperf{0.4}   & 0.9812  \decreasedperf{0.0013} \\
          & Earthformer &  38.9  \decreasedperf{0.7}   & 0.9790  \decreasedperf{0.0035} \\
          & MCVD  &   36.4  \decreasedperf{2.9}  &  0.9676  \decreasedperf{0.0149} \\
          
    \midrule
    \multirow{5}[2]{*}{\rotatebox[origin=c]{90}{Cityscapes}} 
          & Ours      &  27.4   & 0.8874 \\
          & TAU       &  26.4 \decreasedperf{1.0}  & 0.8660 \decreasedperf{0.0214} \\
          & SimVP     &    26.5  \decreasedperf{0.9}    & 0.8717 \decreasedperf{0.0157} \\
          & SimVPv2   &  26.7 \decreasedperf{0.7}     & 0.8738 \decreasedperf{0.0136} \\
          & MCVD      &  19.1 \decreasedperf{8.3}   & 0.8165 \decreasedperf{0.0709} \\
    \bottomrule
    \end{tabular}%
  \label{tab:single_task_baseline}%
  \vspace{-0.5cm}
\end{table}%

\subsection{Main Results} 
\noindent \textbf{Multi-discipline Results. } Tab.~\ref{tab:main_results} shows multi-discipline results of the proposed UniSTD method and other baselines including TAU~\cite{tan2023temporal}, SimVP~\cite{gao2022simvp}, SimVPv2~\cite{tan2022simvp}, and EarthFormer~\cite{gao2022earthformer} across diverse spatiotemporal prediction tasks. 
The results are reported in terms of PSNR, SSIM, CSI, and NINO, where higher values indicate better performance. 

For the TaxiBJ dataset, the proposed UnSTD method achieves a PSNR of 39.6 and an SSIM of 0.98, significantly outperforming all baseline models. The next-best-performing model, SimVP, achieves a PSNR of 29.7 and an SSIM of 0.75. This represents a substantial improvement of approximately 33\% in PSNR and 30\% in SSIM, highlighting the ability of UnSTD to model spatiotemporal dependencies effectively in traffic control scenarios. Similarly, for the Traffic4Cast dataset, UnSTD achieves a PSNR of 30.6 and an SSIM of 0.92. 
In the BAIR and Human3.1M datasets, which are pivotal for trajectory prediction and robot action planning, UnSTD again demonstrates outstanding performance. For the BAIR dataset, UnSTD attains a PSNR of 20.3 and an SSIM of 0.86, outperforming all other models. The best-performing baseline, Earthformer, achieves a PSNR of 16.3 and an SSIM of 0.61, highlighting an improvement of 25\% in PSNR and 41\% in SSIM. Similarly, for Human3.1M, UnSTD achieves a PSNR of 33.2 and an SSIM of 0.98, significantly surpassing the performance of the next-best model, TAU$_{\textbf{3 Tasks}}$, which achieves a PSNR of 23.7 and an SSIM of 0.89.
In driving scene prediction tasks, UnSTD also achieves remarkable results. For example, UnSTD delivers a PSNR of 27.4 and an SSIM of 0.89 on Cityscapes, far outperforming the best-performing baseline, TAU$_{\textbf{3 Tasks}}$, which achieves a PSNR of 19.6 and an SSIM of 0.59. This reflects the UniSTD's capability to understand complex spatiotemporal interactions in urban driving scenarios with unified modeling. On the KITTI dataset, UnSTD achieves a PSNR of 17.2 and an SSIM of 0.61, surpassing the next-best baseline Earthformer. 
For the weather forecasting tasks, UnSTD demonstrates consistently better results, \eg, achieving a CSI of 0.41 while the baselines have only about $0.3$ on SEVIR.

\begin{figure}[t]
  \centering
\includegraphics[width=1.0\linewidth]{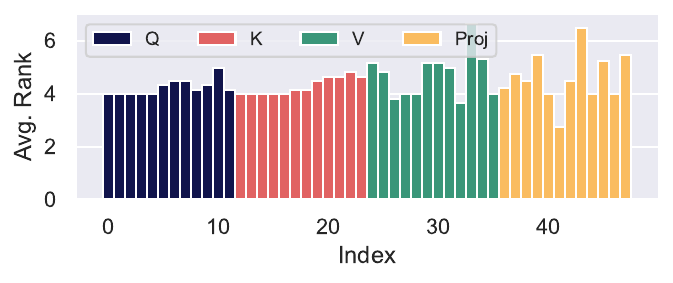}
\vspace{-0.6cm}
\caption{Visualization of the average rank across selected layers (Q, K, V and Proj) in MoEs.}
  \label{fig:exp_vis_rank}
  \vspace{-0.5cm}
\end{figure}

\begin{figure*}[t]
  \centering
\includegraphics[width=1.0\linewidth]{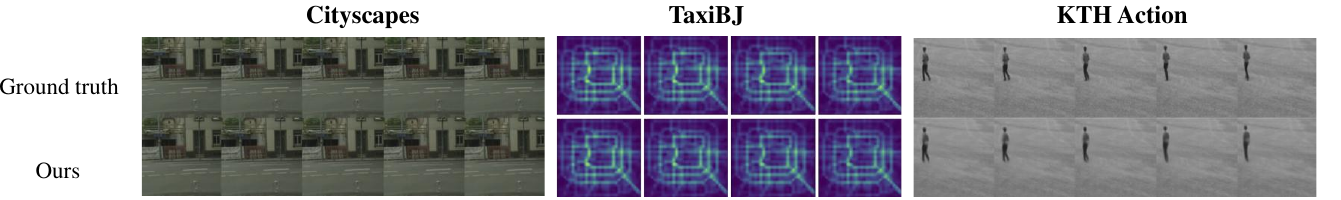}
\caption{Visualization of the prediction results using a shared model.}
\label{fig:vis_prediction_results}
\vspace{-0.25cm}
\end{figure*}

\noindent \textbf{Compared to Single-task Baseline.}
The task-wise comparison in Tab~\ref{tab:single_task_baseline} highlights the superior performance of our unified model on TaxiBJ and Cityscapes datasets. On TaxiBJ, our model achieves the highest PSNR (39.6) and SSIM (0.9825), significantly outperforming MCVD, which lags with a PSNR of 36.4 and an SSIM of 0.9676. On Cityscapes, our model also leads with a PSNR of 27.4 and an SSIM of 0.8874, showing a notable margin over MCVD's PSNR of 19.1 and SSIM of 0.8165. These results emphasize the effectiveness of our unified approach, showing that \emph{joint training across multiple tasks could provide cross-task learning benefits}. 

Furthermore, we have observed the inconsistent performance of existing methods across various tasks, \ie, the TAU is the best baseline for TaxiBJ but not for Cityscapes, and Earthformer, the model designed for Earth system forecasting performs second worst for Cityscapes. These observations reveal that the \emph{existing works heavily rely on task-specific knowledge for designing the architecture which leads to limited generality and cannot be directly applied to other tasks.  }

\subsection{Visualization} 
\noindent\textbf{Rank-adaptive MoE.} Fig.~\ref{fig:exp_vis_rank} illustrates the average rank of MoEs across selected dense layers ($\mathbf{Q}$, $\mathbf{K}$, $\mathbf{V}$, and Proj), where the initial rank of each layer is set to $4.5$. Notably, $\mathbf{Q}$ and $\mathbf{K}$ exhibit stable rank assignments, with earlier layers predominantly ranked around 4, while later layers receive higher ranks. In contrast, $\mathbf{V}$ and Proj display markedly different patterns, with each layer assigned distinct ranks. This aligns with the observations in Fig.~\ref{fig:qkv_norm}, where the norm of $\mathbf{V}$ is more pronounced and distinctive. \\
\noindent\textbf{Model Prediction.}
Fig.~\ref{fig:vis_prediction_results} shows the visualization of predicted frames using our once-for-all model. 
Despite very large differences between inputs (such as TaxiBJ for traffic control and Cityscapes for autonomous driving), the models still achieve accurate results, which further demonstrates the effectiveness of our proposed method.

\subsection{Ablation Study} 

\begin{table}[t]
    \footnotesize
  \centering
  \caption{Effectiveness of proposed Rank-adaptive MoE (AdaMoE) scheme and temporal attention (TAtn.) technique. To save costs, we train the model with 4 tasks. We report the PSNR (higher is better) here. }
  \setlength\tabcolsep{5pt} 
    \begin{tabular}{cccccc}

    \toprule
    AdaMoE   & TAtn. & BAIR  & KTH   & Cityscapes  & TaxiBJ \\
    \midrule
    \cmark     & \cmark     & 20.2    & 28.2    & 27.4    & 39.5 \\
    \cmark     & \xmark     & 20.1 \decreasedperf{0.1}    & 27.7 \decreasedperf{0.5}   & 26.7 \decreasedperf{0.7}   & 39.4 \decreasedperf{0.1} \\ 
    \xmark     & \xmark     & 20.1 \decreasedperf{0.1}   & 27.3  \decreasedperf{0.9}  & 26.2 \decreasedperf{1.2}   & 39.2 \decreasedperf{0.3} \\
    \bottomrule
    \end{tabular}%
    
  \label{tab:effectiveness_proposed_method}%
  \vspace{-0.55cm}
\end{table}%

\noindent\textbf{Effectiveness of Proposed Method.}
Tab.~\ref{tab:effectiveness_proposed_method} evaluates the effectiveness of the proposed Rank-adaptive Mixture-of-Experts (AdaMoE) and Temporal Attention (Temp Attn.) techniques using PSNR (higher is better) across four tasks. 
The configuration integrating both AdaMoE and Temp Attn. consistently achieves the highest PSNR, highlighting their combined efficacy. For instance, on the KTH dataset, the full model achieves a PSNR of 28.2, surpassing configurations without Temp Attn. (27.7) or without both techniques (27.3). Disabling Temp Attn. while retaining AdaMoE results in a moderate decline, with the Cityscapes PSNR dropping from 27.4 to 26.7. The baseline model, devoid of both techniques, performs the worst, with PSNR values such as 39.2 on TaxiBJ compared to 39.5 for the full configuration. These results underscore AdaMoE's role in adaptively managing model complexity and Temp Attn.'s capability to capture temporal dependencies, demonstrating their complementary contributions to enhanced model performance. 
\\
\noindent\textbf{Performance of initialization scheme.} 
In Tab.~\ref{tab:init_method}, we show the advantages of pre-trained (\ie, advantages of the task-agnostic pretraining) initialization schemes across three datasets. 
For BAIR, both LAION~\cite{schuhmann2021laion} and CLIP~\cite{radford2021learning} offer slight improvements over random initialization. 
On the KTH dataset, pre-trained schemes show more significant benefits, with CLIP achieving the best results (PSNR of 28.2, SSIM of 0.92) compared to random initialization (27.1, 0.90). Similarly, for Cityscapes, CLIP achieves the highest performance (PSNR of 27.4, SSIM of 0.89) over random (26.7, 0.87). These results demonstrate that pre-trained models, particularly CLIP, provide a more robust starting point for spatiotemporal tasks. 
In this paper, we choose CLIP weight as our initialization scheme. 

\begin{table}[t]
  \centering
      \footnotesize
  \caption{Performance of initialization scheme. }
    \begin{tabular}{lllllll}
    \toprule
    \multicolumn{1}{c}{\multirow{2}[4]{*}{Init. Scheme}} & \multicolumn{2}{c}{BAIR} & \multicolumn{2}{c}{KTH} & \multicolumn{2}{c}{ Cityscapes } \\
\cmidrule{2-7}                  & PSRN  & SSIM  & PSRN  & SSIM  & PSRN  & SSIM \\
\cmidrule{2-7}    Random & 20.1    & 0.85    & 27.1    & 0.90    & 26.7    & 0.87 \\
    LAION~\cite{schuhmann2021laion}                      & 20.2    & 0.86    & 28.1    & 0.92    & 27.1    & 0.88 \\
    CLIP  ~\cite{radford2021learning}                           & 20.2     & 0.86   & 28.2    & 0.92    & 27.4    & 0.89 \\
    \bottomrule
    \end{tabular}%
  \label{tab:init_method}%
  \vspace{-0.45cm}
\end{table}%

\section{Conclusion}
We address the limitations of existing spatiotemporal models by proposing a unified framework and spatiotemporal decoupling modeling. 
UniSTD leverages a standard Transformer backbone with task-agnostic pretraining and task-specific fine-tuning, ensuring cross-task generality. The decoupling modeling introduces a rank-adaptive mixture-of-expert mechanism and a lightweight temporal module to handle spatial and temporal dependencies. We hope these contributions can advance the field toward general-purpose spatio-temporal learning, reducing reliance on task-specific designs while enhancing adaptability and efficiency.

\noindent \textbf{Acknowledgements.} This work is supported partially by the JC STEM Lab of AI for Science and Engineering, funded by The Hong Kong Jockey Club Charities Trust, the Research Grants Council of Hong Kong (Project No.~CUHK14213224), the National Natural Science Foundation of China~(Grant~No.~62306261), and The Shun Hing Institute of Advanced Engineering (SHIAE) (Grant~No.~8115074).

\balance
{
    \small
    \bibliographystyle{ieeenat_fullname}
    \bibliography{main}
}

\clearpage
\setcounter{page}{1}

\maketitlesupplementary
% \onecolumn
% \newenvironment{links}{%
%   \newcommand{\link}[2]{\par\textbf{##1} --- \url{##2}}%
%   \setlength{\hangindent}{10pt}%
%   \setlength{\parskip}{2pt}%
%   \begin{flushleft}%
% }{%
%   \end{flushleft}%
%   \vskip 1ex%
% }

\setcounter{section}{0}
\section{Implementation Details} 
\noindent \textbf{Architecture. } More details about the architectures of UniSTD are shown in Tab.~\ref{tab:details}. 
The decoder blocks are the same as the Encoder blocks but with the \texttt{TransposedConv2d} layers to replace the \texttt{Conv2d} layers in the Encoder layers for upsampling the features. \\ 
\noindent \textbf{Training Details. } We train the model for 90 epochs with the AdamW optimizer for the weights and SGD optimizer for the trainable rank of MoE. For the Adam optimizer, the weight decay is set to $1e-5$, the learning rate is set to $7e-4$ with cosine learning rate scheduler and the first 5 epochs are used for warm-up (using $1e-8$ learning rate). For the SGD optimizer, we enable the Nesterov momentum and set the learning rate to $0.05$.

\setcounter{section}{1}
\section{Additional Results } 
\begin{table}[h]
    \centering
    \begin{tabular}{c|lcc}
    \toprule
    \multicolumn{1}{c}{Task} & Model & \multicolumn{1}{c}{\makecell{PSNR \\ ($\uparrow$)}} & \multicolumn{1}{c}{\makecell{SSIM \\ ($\uparrow$)}} \\
    \midrule
    \multirow{4}[2]{*}{\rotatebox[origin=c]{90}{BAIR}} 
    & Ours &   20.3    & 0.86 \\
          & TAU   &   19.8    & 0.86 \\
          & SimVPV1 &  20.3     &  0.86  \\
          & SimVPv2 &  19.9    & 0.85 \\
    \midrule
    \multirow{4}[2]{*}{\rotatebox[origin=c]{90}{KTH}} 
          & Ours      &  28.4   & 0.92 \\
          & TAU       &  27.9  & 0.90 \\
          & SimVPv1     &    27.7     & 0.90 \\
          & SimVPv2   &  27.9     & 0.90 \\
    \midrule
    \multirow{4}[2]{*}{\rotatebox[origin=c]{90}{MMNIST}} 
          & Ours      &  20.5   & 0.90 \\
          & TAU       &  18.9  & 0.85 \\
          & SimVPv1     &  19.5    & 0.88 \\
          & SimVPv2   &  19.0     & 0.85 \\

    \bottomrule
    % \caption{Task-wise comparison of our unified model and the single task baselines. }
    \end{tabular}%
    \caption{Task-wise comparison of our unified model and the single task baselines. }
    \label{tab:add_single_task_baseline}
\end{table}
\begin{table}[h]
    \centering
\begin{tabular}{ccc}
    \toprule
    \multicolumn{1}{c}{Task} & Method & \makecell{Trainable Params. \\ ($\downarrow$) } \\
    \midrule
    \multirow{4}[2]{*}{KITTI+Traffic.+KTH} & SimVPv1 & 45.8M \\
          & SimVPv2 & 35.3M \\
          & TAU   & 33.9M \\
          & UniSTD (Ours) & 18.5M \\
          \midrule
    \multirow{4}[2]{*}{\makecell{BAIR+TaxiBJ+Human\\+City.+KTH+Traffic.\\+KITTI}} & SimVPv1 & 61.7M \\
          & SimVPv2 & 47.8M \\
          & TAU   & 45.8M \\
          & UniSTD (Ours) & 23.6M \\
    \bottomrule
    \label{tab:traninable_params}
    % \caption{Number of trainable Parameters of UniT and baseline. }
    \end{tabular}%
    \caption{Number of trainable Parameters of UniT and baseline. }
    \label{tab:traninable_params}
\end{table}

\noindent \textbf{More Evaluation Metrics. } We provide an additional evaluation metric RMSE in Tab.~\ref{tab:rmse}, one can see that our method still yield best performance across various methods.  . \\
\noindent \textbf{Task-wise Training Results. } In Tab.~\ref{tab:add_single_task_baseline}, we show the additional results of our joint trained model compared to the single-task training (independent training) of baselines.  Our model shows significant improvements on metrics of both PSNR and SSIM, this further indicates that the joint training can benefit the learning process of each task. We train the baseline using our training settings for fair comparison. \\ 
\noindent \textbf{Efficiency Analysis. } In Tab.~\ref{tab:flops} and Tab.~\ref{tab:traninable_params}, we show the computational complexity (FLOPs) and number of trainable parameters of the proposed method and baseline, respectively. 
On the one hand, long-range spatial modeling of Transformer allows UniSTD to use much smaller spatial dimensions, thus more efficient on mid/large resolution tasks (e.g., SEVIR, Human, etc) in terms of FLOPs. 
On the other hand, our method uses only about 50\% trainable parameters of the baselines while achieving much better performance.

\begin{table*}[t]
  \centering
  \tiny 
  \caption{Architecture details of UniSTD. }
    \vspace{-0.35cm}
  \setlength\tabcolsep{3pt} 
    \begin{tabular}{p{10.57em}cccccccccc}
    \toprule
    \multicolumn{1}{c}{} & \multicolumn{1}{c}{TaxiBJ} & \multicolumn{1}{c}{Traffic4Cast} & \multicolumn{1}{c}{MMNIST} & \multicolumn{1}{c}{BAIR} & \multicolumn{1}{c}{Human3.1M} & \multicolumn{1}{c}{KTH} & \multicolumn{1}{c}{Cityscapes} & \multicolumn{1}{c}{KITTI} & \multicolumn{1}{c}{SEVIR} & \multicolumn{1}{c}{ENSO} \\
    \midrule
    \multicolumn{1}{c}{GFLOPs} & \multicolumn{1}{c}{6.91} & \multicolumn{1}{c}{26.43} & \multicolumn{1}{c}{7.36} & \multicolumn{1}{c}{139.97} & \multicolumn{1}{c}{31.53} & \multicolumn{1}{c}{29.89} & \multicolumn{1}{c}{85.01} & \multicolumn{1}{c}{37.78} & \multicolumn{1}{c}{88.76} & \multicolumn{1}{c}{62.97} \\
    \midrule
    \multicolumn{1}{l}{Shape: (B*T, C, H, W)} & \multicolumn{1}{l}{(B*4, 2, 32, 32)} & \multicolumn{1}{l}{(B*9, 8, 128,112)} & \multicolumn{1}{l}{(B*10, 1, 64, 64)} & \multicolumn{1}{l}{(B*2, 3, 64, 64)} & \multicolumn{1}{l}{(B*4, 3, 256, 256)} & \multicolumn{1}{l}{(B*10, 1,128, 128)} & \multicolumn{1}{l}{(B*2, 3, 128, 128)} & \multicolumn{1}{l}{(B*10, 3, 128, 160)} & \multicolumn{1}{l}{(B*13, 1, 384, 384)} & \multicolumn{1}{l}{(B*12, 1, 24, 48)} \\
    \midrule
    \textbf{Encoder Block}: \\ \texttt{- Conv2d} \newline{} (stride=1, channels=T') \\ \texttt{- GroupNorm} \\ \texttt{- SiLU} \newline{}\texttt{- Conv2d} \newline{} (stride=2, channels=C')\newline{}\texttt{- GroupNorm}\newline{}{}\texttt{- SiLU} & \multicolumn{1}{c}{*2} & \multicolumn{1}{c}{*3} & \multicolumn{1}{c}{*3} & \multicolumn{1}{c}{*2} & \multicolumn{1}{c}{*4} & \multicolumn{1}{c}{*3} & \multicolumn{1}{c}{*3} & \multicolumn{1}{c}{*3} & \multicolumn{1}{c}{*4} & \multicolumn{1}{c}{*1} \\
    \midrule
    \multicolumn{11}{c}{C'=64} \\
    \midrule
    \multicolumn{1}{c}{Shape: (B, T*C', H', W')} & \multicolumn{1}{c}{(B, 256, 8, 8)} & \multicolumn{1}{c}{(B, 576, 16, 14)} & \multicolumn{1}{c}{(B, 640, 8, 8)} & \multicolumn{1}{c}{(B, 128, 16, 16)} & \multicolumn{1}{c}{(B, 256, 16, 16)} & \multicolumn{1}{c}{(B, 640, 16, 16)} & \multicolumn{1}{c}{(B, 128, 16, 16)} & \multicolumn{1}{c}{(B, 640, 16, 20)} & \multicolumn{1}{c}{(B, 832, 24, 24)} & \multicolumn{1}{c}{(B, 768, 12, 24)} \\
    \midrule
    \textbf{Projection Layer}: \\ \texttt{- Conv2d} \newline{} (stride=1, channels=768) & \multicolumn{1}{c}{*1} & \multicolumn{1}{c}{*1} & \multicolumn{1}{c}{*1} & \multicolumn{1}{c}{*1} & \multicolumn{1}{c}{*1} & \multicolumn{1}{c}{*1} & \multicolumn{1}{c}{*1} & \multicolumn{1}{c}{*1} & \multicolumn{1}{c}{*1} & \multicolumn{1}{c}{*1} \\
    \midrule
    \multicolumn{1}{l}{Shape: (B, 768, H', W')} & \multicolumn{10}{c}{(B, 768, H', W')} \\
    \midrule
     \texttt{- Reshape}  & \multicolumn{10}{c}{N=H'*W', L=768} \\
    \midrule
    \multicolumn{1}{l}{Shape: (B, N, L)} & \multicolumn{1}{c}{(B, 64, 768)} & \multicolumn{1}{c}{(B, 224, 768)} & \multicolumn{1}{c}{(B, 64, 768)} & \multicolumn{1}{c}{(B, 256, 768)} & \multicolumn{1}{c}{(B, 256, 768)} & \multicolumn{1}{c}{(B, 256, 768)} & \multicolumn{1}{c}{(B, 256, 768)} & \multicolumn{1}{c}{(B, 320, 768)} & \multicolumn{1}{c}{(B, 576, 768)} & \multicolumn{1}{c}{(B, 288, 768)} \\
    \midrule
    \textbf{Backbone}: \\ \texttt{- Transformer Blocks (shared)} \newline{}with Rank-Adaptive MoE and  Temp. Attn. & \multicolumn{10}{c}{* 12} \\
    \bottomrule
    \end{tabular}%
  \label{tab:details}%
  \vspace{-0.1cm}
\end{table*}%
\begin{table*}[t]
  \centering
  % \tiny 
  \vspace{-6cm}
  \caption{GFLOPs (lower is better) comparison. }
  \setlength\tabcolsep{3pt} 
    \begin{tabular}{p{10.57em}cccccccccc}
    \toprule
    \multicolumn{1}{c}{} & \multicolumn{1}{c}{TaxiBJ} & \multicolumn{1}{c}{Traffic4Cast} & \multicolumn{1}{c}{MMNIST} & \multicolumn{1}{c}{BAIR} & \multicolumn{1}{c}{Human3.1M} & \multicolumn{1}{c}{KTH} & \multicolumn{1}{c}{Cityscapes} & \multicolumn{1}{c}{KITTI} & \multicolumn{1}{c}{SEVIR} & \multicolumn{1}{c}{ENSO} \\
    \midrule
    \multicolumn{1}{c}{UniSTD (Ours)} & \multicolumn{1}{c}{6.91} & \multicolumn{1}{c}{\textbf{26.43}} & \multicolumn{1}{c}{\textbf{7.36}} & \multicolumn{1}{c}{\textbf{139.97}} & \multicolumn{1}{c}{\textbf{31.53}} & \multicolumn{1}{c}{\textbf{29.89}} & \multicolumn{1}{c}{\underline{85.01}} & \multicolumn{1}{c}{\textbf{37.78}} & \multicolumn{1}{c}{\textbf{88.76}} & \multicolumn{1}{c}{62.97} \\
    \midrule
    \multicolumn{1}{c}{SimVP$_{\text{v1}}$} & \multicolumn{1}{c}{3.53} & \multicolumn{1}{c}{40.01} & \multicolumn{1}{c}{15.15} & \multicolumn{1}{c}{277.64} & \multicolumn{1}{c}{231.80} & \multicolumn{1}{c}{\underline{50.06}} & \multicolumn{1}{c}{127.70} & \multicolumn{1}{c}{\underline{62.60}} & \multicolumn{1}{c}{426.92} & \multicolumn{1}{c}{25.58} \\
    \midrule
    \multicolumn{1}{c}{SimVP$_{\text{v2}}$} & \multicolumn{1}{c}{\underline{2.54}} & \multicolumn{1}{c}{32.06} & \multicolumn{1}{c}{12.28} & \multicolumn{1}{c}{194.77} & \multicolumn{1}{c}{166.59} & \multicolumn{1}{c}{64.26} & \multicolumn{1}{c}{88.79} & \multicolumn{1}{c}{80.35} & \multicolumn{1}{c}{398.34} & \multicolumn{1}{c}{22.82} \\
    \midrule
    \multicolumn{1}{c}{TAU} & \multicolumn{1}{c}{\textbf{2.43}} & \multicolumn{1}{c}{\underline{30.62}} & \multicolumn{1}{c}{\underline{11.75}} & \multicolumn{1}{c}{\underline{185.44}} & \multicolumn{1}{c}{\underline{158.83}} & \multicolumn{1}{c}{61.25} & \multicolumn{1}{c}{\textbf{84.78}} & \multicolumn{1}{c}{76.59} & \multicolumn{1}{c}{\underline{379.91}} & \multicolumn{1}{c}{21.79} \\
    \bottomrule
    \end{tabular}%
  \label{tab:flops}%
  % \vspace{-0.1cm}
\end{table*}%

\begin{table*}[t]
  \centering
  % \tiny 
    \vspace{-6cm}
  \caption{{RMSE ($\downarrow$) metrics. }}
    \vspace{-0.35cm}
  \setlength\tabcolsep{3pt} 
    \begin{tabular}{p{10.57em}cccccccccc}
    \toprule 
    \multicolumn{1}{c}{} & \multicolumn{1}{c}{TaxiBJ} & \multicolumn{1}{c}{Traffic4Cast} & \multicolumn{1}{c}{MMNIST} & \multicolumn{1}{c}{BAIR} & \multicolumn{1}{c}{Human3.1M} & \multicolumn{1}{c}{KTH} & \multicolumn{1}{c}{Cityscapes} & \multicolumn{1}{c}{KITTI} \\
    \midrule
    \multicolumn{1}{c}{UniSTD \textbf{(Ours)}} & \multicolumn{1}{c}{0.54} & \multicolumn{1}{c}{8.78} & \multicolumn{1}{c}{6.27} & \multicolumn{1}{c}{11.54} & \multicolumn{1}{c}{10.46} & \multicolumn{1}{c}{5.42} & \multicolumn{1}{c}{10.82} & \multicolumn{1}{c}{39.49}  \\
    \midrule
    \multicolumn{1}{c}{SimVP$_{\text{3 task}}$} & \multicolumn{1}{c}{2.86} & \multicolumn{1}{c}{-} & \multicolumn{1}{c}{-} & \multicolumn{1}{c}{18.79} & \multicolumn{1}{c}{35.45} & \multicolumn{1}{c}{\underline{-}} & \multicolumn{1}{c}{-} & \multicolumn{1}{c}{\underline{-}}  \\
    \midrule
    \multicolumn{1}{c}{TAU$_{\text{3 task}}$} & \multicolumn{1}{c}{1.69} & \multicolumn{1}{c}{\underline{-}} & \multicolumn{1}{c}{\underline{-}} & \multicolumn{1}{c}{\underline{-}} & \multicolumn{1}{c}{34.48} & \multicolumn{1}{c}{-} & \multicolumn{1}{c}{26.56} & \multicolumn{1}{c}{76.59}  \\
    \midrule
    \multicolumn{1}{c}{UniST~\cite{yuan2024unist}} & \multicolumn{1}{c}{0.90} & \multicolumn{1}{c}{-} & \multicolumn{1}{c}{-} & \multicolumn{1}{c}{25.76} & \multicolumn{1}{c}{-} & \multicolumn{1}{c}{23.36} & \multicolumn{1}{c}{-} & \multicolumn{1}{c}{56.95}  \\
    \bottomrule
    \end{tabular}%
  \label{tab:rmse}%
  \vspace{-0.4cm}
\end{table*}%

\end{document}